\documentclass[11pt]{article}
\usepackage{coling2020}
\usepackage{times}
\usepackage{url}
\usepackage{latexsym}

\usepackage{colortbl}

\colingfinalcopy 

\title{Uralic Language Identification (ULI) 2020 shared task dataset and the Wanca 2017 corpus}
  
  \author{Tommi Jauhiainen \\
  Department of Digital Humanities \\
  University of Helsinki \\
  {\tt tommi.jauhiainen@helsinki.fi  } \\\And
  Heidi Jauhiainen \\
  Department of Digital Humanities \\
  University of Helsinki \\
  {  \tt heidi.jauhiainen@helsinki.fi} \\\AND
  Niko Partanen \\
  Department of Finnish, Finno-Ugrian\\and Scandinavian Studies \\
  University of Helsinki \\
  {\tt niko.partanen@helsinki.fi} \\\And
  Krister Lind\'en \\
  Department of Digital Humanities \\
  University of Helsinki \\
  {\tt krister.linden@helsinki.fi} \\}

\date{}

\begin{document}
\maketitle
\begin{abstract}
This article introduces the Wanca 2017 corpus of texts crawled from the internet from which the sentences in rare Uralic languages for the use of the Uralic Language Identification (ULI) 2020 shared task were collected. We describe the ULI dataset and how it was constructed using the Wanca 2017 corpus and texts in different languages from the Leipzig corpora collection. We also provide baseline language identification experiments conducted using the ULI 2020 dataset.
\end{abstract}

\section{Introduction}
\label{intro}

As part of the Finno-Ugric and the Internet project (SUKI), we have collected textual material for some of the more endangered Uralic languages from the internet \cite{jauhiainen5}. In this paper, we introduce the Wanca 2017 corpus which will be published in the Language Bank of Finland\footnote{\url{https://www.kielipankki.fi/language-bank/}} as a downloadable package as well as through the Korp\footnote{\url{https://korp.csc.fi}} concordance service. We used the earlier version of the corpus, Wanca 2016 \cite{Jauhiainen-Jauhiainen-JauhiainenRDHum19}, together with corpora available from the Leipzig corpora collection \cite{goldhahn2012building} to create a training dataset for the ULI 2020 shared task. The ULI 2020 shared task was organized as part of the VarDial 2020 Evaluation campaign. We also performed a baseline language identification experiment for the ULI dataset using the HeLI method described by \newcite{jauhiainen6}.

In this paper, we first introduce some related work and resources for language identification and the Uralic languages in Section 2. We then describe the Wanca 2017 corpus and its creation in Section 3. In Section 4, we give a detailed description of the creation of the dataset for the ULI 2020 shared task as well as the information about the baseline language identification experiments using the corpus, and provide some error analysis for the results of those experiments.


\section{Related work}

In this section, we first introduce some previous work on language identification of texts, then we give a short introduction to the Uralic languages and present some of the text corpora already available for those languages.

\subsection{Language identification in texts}

In this paper, we focus on language identification in texts as opposed to language identification in speech. By language identification, we mean the labeling of sentences or texts by language labels from a given label set, which is the test set-up in the ULI shared task. By defining the problem this way, we have ignored two challenges in language identification: detection of unknown languages and multilingual texts. In unknown language detection, the language identifier can be presented with texts that are written in a language that it has not been trained in. Multilingual texts are simply texts containing more than one written language. Actually, in the strict sense, some of the sentences in the training and test sets of the ULI task can be considered multilingual as they may include some words in languages other than the main language of the sentence. In the ULI task, the target is however, to simply label the main language for each sentence.

A recent survey concerning language identification in texts by \newcite{jauhiainen2019b} gives a thorough introduction to the subject.

\subsection{Uralic languages}

In this section we provide a general overview to the Uralic language family, with specific attention to development of the written standards and contemporary use, as this is closely connected to the resources available for the language identification task. 
The Uralic language family contains 30-40 languages, and shows considerable diversity at all levels. 
Handbooks about the family include \newcite{abondolo1998a} and \newcite{sinor1988a}, and new handbooks are currently under preparation \cite{laaksoEtAl2020a,abondolo2021}. 
The Uralic language family is one of the most reliably established old language families in the world, and can be compared with the Indo-European language family in its time depth and variation, although the exact dating of the family is a matter of on-going research. 

Geographically the Uralic languages are spoken in Northern Eurasia, with the Saami languages in the Scandinavia representing the westernmost extent, and the Nganasans at the Taimyr Peninsula are the easternmost Uralic language speakers. 
In the south, Hungarian, a geographical outlier, is spoken in the Central European Carpathian Basin. 
The majority of the Uralic languages are spoken within the Russian Federation. 
The wide geographical area also has resulted in different subsistence strategies and livelihoods, historically, and also in various contemporary conditions. 
Only three Uralic languages, Estonian, Finnish and Hungarian, are spoken as the majority language of a country. 
These languages are not endangered, but they have closely related varieties that often are, as are all other Uralic languages. 

Some Uralic languages are also already extinct. 
This is the case with Kemi Saami, which ceased to be spoken in the 19th century, and Kamas, the last speaker of which died in 1989. 
The former is represented in this shared task. 
Still spoken Uralic languages form a continuum also in their number of speakers, as the smallest languages, such as Inari Saami and Skolt Saami, have only hundreds of speakers, and Nganasan maybe slightly more than one hundred \cite[17]{wagner2018grammar}. 
In contrary, languages such as Mari or Udmurt have hundreds of thousands of speakers, and are used actively in various spheres of modern society. 
They are also endangered due to interrupted intergenerational language transmission and disruption of the traditional speech communities. 
When it comes to the online presence, or generally to available textual representations of these languages, historical developments in their standardization and language planning are in a very central role. 
This was largely outlined by Soviet language policy, described in detail in \newcite{grenoble2003language}. 
It has also been typical for the Uralic languages spoken in Russia that the orthographies have changed numerous times. 
\newcite{siegl2015uneven} discuss four case studies about the possible variation in the degrees of contemporary literacy and development of the written standards. 
There are numerous languages in the Wanca corpus for which the ortographies were developed in the late 19th or early 20th century. 
This pertains especially to many languages spoken in the Soviet Union, including Ingrian, Karelian, Livvi-Karelian, Vepsian, Komi-Permyak, Komi-Zyrian, Udmurt, Khanty, Mansi and Tundra Nenets. 
Even with very closely related languages, such as two Komi written standards, two Mari written standards or two Mordva written standards, the contemporary orthographies and the varieties themselves contain numerous differences in their phonology and spelling conventions that make distinguishing the language of a text almost always straightforward, at least to a specialist. 
These differences are large enough that from the perspective of computational linguistics, distinct infrastructure usually has to be developed for each variety, even the actual linguistic differences would be minor. 
For an example about challenges in creating Komi-Permyak and Komi-Zyrian infrastructure, see \newcite{rueter2020a}. 

Some of these orthographies were more successful than others, and there is large variation in when exactly the currently used system was established, and what level of stability they have. 
This was also the case for Nganasan, but the orthography created in 1986 was never widely used, and in the current orthography the conventions vary with author and editor \cite{wagner2018grammar}. 
For languages such as Votic, the current orthography was developed only in the 2000s \cite[3]{ernits2006a}. 
Another good example is Tundra Nenets, which has had the current orthography since the 1940s, and which has all in all 100 titles published. 
The language is also partially used in local newspapers. \cite{nikolaeva2014grammar}. 
The small number of sentences in this dataset probably indicates, however, that the online visibility is relatively small. 
At the same time a relatively small Sami language, Skolt Sami with approximately 300 speakers, is represented in the dataset by thousands of sentences. 
The Skolt Sami orthography was developed in the 1970s and the knowledge of the writing standard is not complete in the whole community \cite[26,37]{feist2015a}, but the language has been officially recognized in Finland and has received support, which may explain why it appears to have more online presence than some other languages of the same size. 

Thereby the contemporary online presence of these languages is a complex combination of many historical factors. 
However, we can generally say that those languages with more widely used and taught orthographies, and with a substantial speaker base, do have enough materials online that crawling up to several million tokens is possible. 
With smaller languages the situation is different and much more varying. 
There is also the aspect of time, as continuous use accumulates increasingly larger resources. 
What it comes to extinct languages, their corpora have to be considered finite. 

\subsection{Corpora for Uralic languages}

For the Uralic languages that are the majority language of a country, that is Finnish, Estonian, and Hungarian, many large text corpora already exist. For example, there is the Suomi 24 Corpus\footnote{\url{http://urn.fi/urn:nbn:fi:lb-2017021506}} with over 250 million Finnish sentences from a social networking website available from the Language Bank of Finland, and the Europarl corpus\footnote{\url{https://www.statmt.org/europarl/}} with over 600,000 sentences of Hungarian and Estonian \cite{Koehn:2005}. The Leipzig Corpora Collection\footnote{\url{https://wortschatz.uni-leipzig.de/en/download}} has texts also for some of the more rare Uralic languages: Eastern Mari, Komi, Komi-Permyak, Northern Sami, Udmurt, Võro, and Western Mari. The Giellatekno research group has three Korp installations for Uralic languages: one\footnote{\url{http://gtweb.uit.no/korp/}} for Saami languages, one\footnote{\url{http://gtweb.uit.no/f_korp/}} for Kven, Meänkieli, Veps, and Võro, and one\footnote{\url{http://gtweb.uit.no/u_korp/}} for Komi-Zyrian, Komi-Permyak, Udmurt, Moksha, Erzya, Hill Mari, and Meadow Mari. The Wanca in Korp corpus contains texts in all the aforementioned languages as well as some additional Uralic languages.\footnote{\url{http://urn.fi/urn:nbn:fi:lb-2019052402}}

\section{Wanca 2017 corpus}

The Wanca 2017 corpus is the product of a re-crawl performed by the SUKI project in October 2017. The target of the re-crawl was to download and check the availability of the then current version of the Wanca service of about 106,000 pages. The crawl managed to download over 70\% of the target urls. We processed the downloaded pages following the strategy presented by \newcite{jauhiainen2020building} as follows.

First, all the text from each of the 78,685 downloaded pages was sent to a language set identification service. We retained only the pages which had at least 2\% text in one of the minority Uralic languages and the one that was most prominent was set as the page language.
 The retained pages contained a total of 1,515,068 lines and along with the lines, the identified language of the original page was kept. The lines were checked for duplicates, which left 446,233 unique lines. If the duplicates came from pages with different identified language, all those languages were set as the previously known language of the line. Each line was then again sent to the language set identifier which was only allowed to consider the previously known minority languages of the line as well as all non-relevant languages. Again only such lines were retained which included at least one relevant language, leaving 356,637 lines. Next, a language independent sentence extraction algorithm was run on each line and 560,821 sentences were extracted with 477,109 unique sentences. After this, one more round of language set identification was performed. Of the minority Uralic languages, the service was again allowed to consider only those in the list of the previously known language of a sentence, but this time the absolute majority language of the identification was set as the language of the sentence. The resulting corpus contains 447,927 sentences in relevant languages divided as shown in the Wanca 2017 column of Table~\ref{UralicLanguages}.

\begin{table}[h]
\renewcommand{\tabcolsep}{1mm}
\renewcommand{\arraystretch}{.75}
\begin{tabular}{lrrrr}
\rowcolor[gray]{.8}  & \textbf{Wanca 2016} & \textbf{ULI 2020 training} & \textbf{Wanca 2017}  & \textbf{ULI 2020 test}        \\
\emph{Finnic}			\\
\rowcolor[gray]{.9}  \hspace{3mm} Estonian, Standard (\textbf{ekk}) & - & 10,000	& -	& 10,000\\
\hspace{3mm} Finnish (\textbf{fin}) & - & 1,000,000	& -	& 10,000\\
\rowcolor[gray]{.9}  \hspace{3mm} \textbf{Finnish, Kven (fkv)} & 2,156 & 2,156 & 1,499 & 23\\
\hspace{3mm} \textbf{Finnish, Tornedalen (fit)} & 5,203 & 5,203 & 4,517 & 100\\
\rowcolor[gray]{.9}  \hspace{3mm} \textbf{Ingrian (izh)} & 81 & 81 & 80 & -\\
\hspace{3mm} \textbf{Karelian (krl)} & 2,593 & 2,593 & 2,513 & 94\\
\rowcolor[gray]{.9}  \hspace{3mm} \textbf{Liv (liv)} & 705 & 705 & 343 & 68\\
\hspace{3mm} \textbf{Livvi-Karelian (olo)} & 9,920 & 9,920 & 6,486 & 179\\
\rowcolor[gray]{.9}  \hspace{3mm} \textbf{Ludian (lud)} & 771 & 771 & 411 & 185\\
\hspace{3mm} \textbf{Veps (vep)} & 13,461 & 13,461 & 9,122 & 2,453\\
\rowcolor[gray]{.9}  \hspace{3mm} \textbf{Vod (vot)} & 20 & 20 & 11 & -\\
\hspace{3mm} \textbf{Võro (vro)} & 66,878 & 66,878 & 61,430 & 443\\
\rowcolor[gray]{.9}  Hungarian (\textbf{hun}) & - & 1,000,000 & - & 10,000			\\
\textbf{Khanty (kca)} & 1,006 & 1,006 & 940 & 24\\
\rowcolor[gray]{.9}  \textbf{Mansi (mns)} & 904 & 904 & 825 & 1\\
\emph{Mari}			\\
\rowcolor[gray]{.9}  \hspace{3mm} \textbf{Mari, Hill (mrj)} & 30,793 & 30,793 & 22,986 & 18\\
\hspace{3mm} \textbf{Mari, Meadow (mhr)} & 110,216 & 110,216 & 38,278 & 3,768\\
\rowcolor[gray]{.9}  \emph{Mordvin} 	& & & &	\\
\hspace{3mm} \textbf{Erzya (myv)} & 28,986 & 28,986 & 16,273 & 1,153\\
\rowcolor[gray]{.9}  \hspace{3mm} \textbf{Moksha (mdf)} & 21,571 & 21,571 & 15,170 & 724\\
\emph{Permian}			\\
\rowcolor[gray]{.9}  \hspace{3mm} \textbf{Komi-Permyak (koi)} & 8,162 & 8,162 & 6,104 & -\\
\hspace{3mm} \textbf{Komi-Zyrian (kpv)} & 21,786 & 21,786 & 18,966 & 254\\
\rowcolor[gray]{.9}  \hspace{3mm} \textbf{Udmurt (udm)} & 56,552 & 56,552 & 42,545 & 3,562\\
\emph{Sami}	& & & &		\\
\rowcolor[gray]{.9}  \hspace{3mm} \textbf{Sami, Inari (smn)} & 15,469 & 15,469 & 14,405 &  228\\
\hspace{3mm} \textbf{Sami, Kemi (sjk)} & 19 & 19 & - & - \\
\rowcolor[gray]{.9}  \hspace{3mm} \textbf{Sami, Kildin (sjd)} & 132 & 132 & 59 & 13\\
\hspace{3mm} \textbf{Sami, Lule (smj)} & 10,605 & 10,605 & 5,644 & 400\\
\rowcolor[gray]{.9}  \hspace{3mm} \textbf{Sami, North (sme)} & 214,226 & 214,226 & 165,009 & 6,009\\
\hspace{3mm} \textbf{Sami, Skolt (sms)} & 7,819 & 7,819 & 6,696 & 202\\
\rowcolor[gray]{.9}  \hspace{3mm} \textbf{Sami, South (sma)} & 15,380 & 15,380 & 7,204 & 355\\
\hspace{3mm} \textbf{Sami, Ume (sju)} & 124 & 124 & 4 & 1\\
\rowcolor[gray]{.9}  \emph{Samoyed}		& & & &			\\
\hspace{3mm} \textbf{Nenets (yrk)} & 443 & 443 & 407 & 58\\
\rowcolor[gray]{.9}  \hspace{3mm} \textbf{Nganasan (nio)} & 62 & 62 & - & -\\
\end{tabular}
\caption{The number of sentences in Uralic languages for each dataset.}
\label{UralicLanguages}
\end{table}

\section{The ULI 2020 shared task}

The ULI 2020 shared task was organized as a part of the VarDial 2020 Evaluation Campaign.\footnote{\url{https://sites.google.com/view/vardial2020/evaluation-campaign}} The evaluation campaign is the 7th incarnation of a series of shared tasks concentrating on close languages which have always incorporated some form of language identification tasks \cite{zampieri6,zampieri8,malmasi9,zampieri9,zampieri12,zampieri13}.

\subsection{The dataset for the shared task}

We had decided to use the Wanca 2016 corpus \cite{Jauhiainen-Jauhiainen-JauhiainenRDHum19} as training material for the task and extract a test set of new sentences from the Wanca 2017 corpus. As Wanca 2017 was not a real web-crawl, but only included downloading links already existing in the Wanca portal, it was in doubt how many completely new sentences the test set would have. For the ULI 2020 test set, we compared the Wanca 2017 corpus with the Wanca 2016 corpus and kept such sentences that were only found on the 2017 edition. This set including 25,547 sentences was then checked by us and we manually removed all doubtful sentences from the test set, concentrating on improving precision over recall. We were left with a total of 20,315 sentences divided between the minority Uralic languages as seen in the "ULI 2020 test" column of the Table~\ref{UralicLanguages}.

In addition to the relevant languages, the test set includes sentences in 149 other languages from the Leipzig Corpora collection \cite{goldhahn2012building}. The three largest Uralic languages have been included in this category. The download links for the training data for these non-relevant languages were distributed by the task organizers only to participating teams. In total, the training data for the task consisted of 63,772,445 sentences in non-relevant and 646,043 sentences in relevant languages, totaling 64,418,488 sentences. The list of the non-relevant languages is available at the Evaluation campaign website.\footnote{\url{https://sites.google.com/view/vardial2020/evaluation-campaign/uli-shared-task}}

\subsection{Three tracks}

The ULI 2020 shared task included three tracks. The tracks were not just about distinguishing between Uralic languages themselves, but also distinguishing the Uralic languages from the 149 non-relevant languages. The training and the test data for each of the tracks was the same and in each track every line in the test set was to be identified. The difference between the tracks was how the resulting scores were calculated, which significantly affects how the used classifying algorithms should be trained.

The first track of the shared task considered all the relevant languages equal in value and the aim was to maximize their average F-score. This is important when one is interested to find also the very rare languages included in the set of relevant languages. The result was the average of the macro-F1-scores of all the 29 relevant languages present in the training set. If the correct number of true positives for a language was zero, then precision was 100\% if no false positives were predicted. If false positives were predicted, the precision was zero. So, for those five languages (Ingrian, Vod, Komi-Permyak, Kemi Sami, and Nganasan) that were part of the training set, but did not appear in the test set, the recall was always 100\% and precision was either 100\% (if no instances of these languages were predicted in the test set) or 0\% (if even one sentence was labeled as one of them).

The second track considered each sentence in the test set that is written in or is predicted to be in a relevant language as equals. When compared with the first track, this track gave less importance to the very rare languages as their precision was not so important when the resulting F-score was calculated due to their smaller number of sentences. The resulting F-score was calculated as a micro-F1 over the sentences in the test set for sentences in the relevant languages as well as those that were predicted to be in relevant languages.

In the first two tracks, there was no difference between the non-relevant languages when the F1-scores were calculated. The third track, however, did not concentrate on the 29 relevant languages, but instead the target was to maximize the average F-score over all the 178 languages present in the training set. This track was the language identification shared task with the largest number of languages to date (The ALTW 2010 shared task organized by \newcite{baldwin1} included 74 languages). The F-score was calculated as a macro-F1 score over all the languages in the training set.

\subsection{Baseline experiments}

The baseline experiments were conducted using a language identifier based on the HeLI method \cite{jauhiainen2}. In the HeLI method each word in the mystery text has equal weight when determining the language of a text. Each word is divided into character \emph{n}-grams, where the maximum length of the character sequences, \(n_{max}\), is determined using a training and development sets. Other tunable parameters include a cut-off, \(c\), for the frequency of features used as well as a penalty value, \(p\), for unseen features. Instead of tuning the parameters using the ULI 2020 training set, we used the parameters presented by \newcite{jauhiainen6}: \(n_{max}=6\), \(c=0.0000005\), and \(p=7\). As did \newcite{jauhiainen6}, we used the relative frequency of features as a cut-off instead of a raw frequency as the training corpora were of very different sizes. Only lowercased alphabetical characters were used in the language models. Due to HeLI using space character to separate words, there was a special 'sanity check' algorithm for texts including more than 50\% CJK (Chinese-Japanese-Korean) characters, which gave all non-CJK languages a high penalty. The HeLI implementation used is almost exactly the same as the "TunnistinPalveluFast" available from GitHub.\footnote{\url{https://github.com/tosaja/TunnistinPalveluFast}}

We did only one common run for all three tracks of the shared task. The results are listed in Table~\ref{ULIbaseline}.

\begin{table}[h]
\begin{center}
\renewcommand{\arraystretch}{.75}
\begin{tabular}{ll}
\rowcolor[gray]{.8}  \textbf{Language} & \textbf{F-score} \\
ULI subtask 1, relevant macro F1 & 0.8004\\
\rowcolor[gray]{.9}  ULI subtask 2, relevant micro F1 & 0.9632\\
ULI subtask 3, macro F1 & 0.9252\\
\end{tabular}
\end{center}
\caption{The number of sentences in Uralic languages for each dataset.}
\label{ULIbaseline}
\end{table}

Table~\ref{ConfusionOne} displays a confusion matrix showing two of the worst performing languages on Track 1: Ingrian and Votic. There were no real instances of Ingrian in the test set, but our baseline-identifier had identified three sentences of Ludian and one sentence of Karelian as Ingrian. These three languages are all closely related, but also one sentence of Sundanese was identified as Ingrian. The sentence in question is "Unggal lempir kawengku ku tilu padalisan." Both words "ku" and "tilu" are found in the Wanca 2016 corpus for Ingrian, which gives a hint of the reason for the mistake. Another language with an F-score of zero was Votic. Two Ludian sentences were identified as Votic, which is again understandable due to the languages being relatives, but also one sentence in Southern Sotho was identified as Votic: "Madinayne a ja dikokwanyana." As it happens, "a" is the most common word in the Wanca 2016 corpus for Votic and "ja" the sixth most common.

\begin{table}[h]
\small
\begin{center}
\renewcommand{\arraystretch}{.75}
\begin{tabular}{lllllllllllll}
\rowcolor[gray]{.8}  \textbf{Language} & fin  & fit  & fkv  & hat & \textbf{izh}  & kpv & krl   & lud  & sot   & sun  & swe & \textbf{vot} \\
Finnish (fin) & \emph{9,931} & 53 & 7 & & & & 3 & 1 & & & 1 & \\
\rowcolor[gray]{.9}  Tornedalen Finnish (fit) & 5 & \emph{91} & 2 & 1 & & & & & & & 1 & \\
Kven (fkv) & & 3 & \emph{19} & & & & & & & & & \\
\rowcolor[gray]{.9}  Haitian (hat) & & & & \emph{9,924} & & & & & & & & \\
\textbf{Ingrian (izh)} & & & & & & & & & & & & \\
\rowcolor[gray]{.9}  Komi-Zyrian (kpv) & & 1 & & & & \emph{246} & & & & & & \\
Karelian (krl) & 1 & & & & \textbf{1} & & \emph{80}& & & & & \\
\rowcolor[gray]{.9}  Ludian (lud) & 2 & & & & \textbf{3} & & & \emph{144} & & & & \textbf{2} \\
Southern Sotho (sot) & & & & & & & & & \emph{9,962} & & & \textbf{1} \\
\rowcolor[gray]{.9}  Sundanese (sun) & & & & 1 & \textbf{1} & & & & 1 & \emph{5,451} & & \\
Swedish (swe) & 1 & & & & & & & & & & \emph{9,981} & \\
\rowcolor[gray]{.9}  \textbf{Votic (vot)} & & & & & & & & & & & & \\
\end{tabular}
\end{center}
\caption{Confusion matrix of some of the worst performing languages on Track 1.}
\label{ConfusionOne}
\end{table}

Table~\ref{ConfusionTwo} shows the languages which were confused with Võro, the worst performing language on Track 2. Võro is an extremely close language to Standard Estonian, both spoken in modern Estonia. None of the sentences in Võro were identified as Standard Estonian, however over a thousand sentences (out of 10,000) in Standard Estonian were identified as Võro. Actually the only mistake identifying sentences in Võro was when the sentence ""Õdaguhe" (film) ; 20:35 . " was identified as Northern Azerbaijani.

\begin{table}[h]
\small
\begin{center}
\renewcommand{\arraystretch}{.75}
\begin{tabular}{lllllllllllll}
\rowcolor[gray]{.8}  \textbf{Language} & azj  & ekk  & fin  & gsw & hif  & ita & lud & sun & tso & vec & \textbf{vro} & wuu \\
N. Azerbaijani (azj) & \emph{9,896} & & & & & & & & & & & \\
\rowcolor[gray]{.9}  Std. Estonian (ekk) & 3 & \emph{8,717} & 16 & 3 & 4 & 6 & 3 & & & 5 & \textbf{1,052} & \\
Finnish (fin) & & & \emph{9,931} & & & & 1 & & & & \textbf{1} & \\
\rowcolor[gray]{.9}  Swiss German (gsw) & & & & \emph{9,409} & 1 & 6 & 1 & & & 5 & \textbf{1} & \\
Fiji Hindi (hif) & & &  1 & 2 & \emph{9,246} & 4 & & & 1 & & \textbf{1} & 1\\
\rowcolor[gray]{.9}  Italian (ita) & & & & 1 & 2 & \emph{8,723} & & & 1 & 1,026 & \textbf{1} & \\
Ludian (lud) & & & 2 & 1 & & & \emph{144} & & & & \textbf{1} & \\
\rowcolor[gray]{.9}  Sundanese (sun) & & & & & 2 & 1 & & \emph{5,451} & 4 & & \textbf{1} & \\
Tsonga (tso) & & & & & & & & & \emph{9,991} & & \textbf{1} & \\
\rowcolor[gray]{.9}  Venetian (vec) & & & & 3 & & 1,296 & & & & \emph{747} & \textbf{1} & \\
\textbf{Võro (vro)} & \textbf{1} & & & & & & & & & & \emph{\textbf{442}} & \\
\rowcolor[gray]{.9}  Wu Chinese (wuu) & 1 & & & 2 & 9 & & & 8 & & & \textbf{1} & \emph{6,103} \\
\end{tabular}
\end{center}
\caption{Confusion matrix for Võro, the worst performing language on Track 2.}
\label{ConfusionTwo}
\end{table}

To illustrate the identification errors in Track 3, we selected some of the worst performing languages and created a confusion matrix which is presented in the Table~\ref{ConfusionThree}. Bashkir and Tatar are closely related Turkic languages spoken in Russia. According to \newcite{tyers2012prototype}, their orthographical system are fairly different, which might indicate that the corpora used could be noisier than average. The extremely closely related languages Bosnian and Croatian have always been a problem for the non-discriminative HeLI method as is evidenced by the poor results in the DSL shared tasks of 2015, 2016, and 2017 \cite{jauhiainen2019c}. Wu Chinese was identified as Mandarin Chinese over 30\% of the time. Character-based methods should be used instead of word-based methods when word-tokenization is a problem and the simple CJK algorithm included in the baseline-identifier just helps to correct some of the problems between CJK and non-CJK languages, but does not help in distinguishing between CJK languages. The trio of close languages Indonesian, Javanese, and Sundanese got confused to the point of Indonesian being more often identified as Sundanese than Indonesian. Low German (nds-nl\_wikipedia\_2016\_10K) was almost never identified as such (nds\_wikipedia\_2010\_100K), but mostly as Limburgan (lim-nl\_web\_2015\_300K). This seems to be due to the writing system of Low German being in flux and the nds.wikipedia\footnote{\url{https://nds.wikipedia.org/wiki/Plattdüütsch}} and nds-nl.wikipedia\footnote{\url{https://nds-nl.wikipedia.org/wiki/Nedersaksisch}} being different entities.

\begin{table}[h]
\small
\begin{center}
\begin{tabular}{llllllllllll}
\rowcolor[gray]{.8}  \textbf{Language} & \textbf{bak}  & \textbf{bos}  & \textbf{cmn}  & \textbf{hrv} & \textbf{ind}  & \textbf{jav}  & \textbf{lim}  & \textbf{nds}   & \textbf{sun}  & \textbf{tat} & \textbf{wuu} \\
Bashkir (bak) & \emph{6,961} & & & & & & & & & 3,037 & \\
\rowcolor[gray]{.9}  Bosnian (bos) & & \emph{4,403} & & 5,593 & & & & & & & \\
Mandarin Chinese (cmn) & & & \emph{9,562} & & & & & & & & 273 \\
\rowcolor[gray]{.9}  Croatian (hrv) & & 1,134 & & \emph{8,864} & & & & & & & \\
Indonesian (ind) & & & & & \emph{3,102} & 14 & & & 4,858 & & \\
\rowcolor[gray]{.9}  Javanese (jav) & & & & & 1,451 & \emph{4,626} & & & 3,619 & & \\
Limburgan (lim) & & & & & & & \emph{9,540} & 10 & & & \\
\rowcolor[gray]{.9}  Low German (nds) & & & & & & & 5,625 & \emph{182} & & & \\
Sundanese (sun) & & & & & 87 & 4,330 & 1 & & \emph{5,451} & & \\
\rowcolor[gray]{.9}  Tatar (tat) & 3,784 & & & & & & & & & \emph{6,215} & \\
Wu Chinese (wuu) & & 1 & 3,610 & & 1 & 2 & 1 & & 8 & 1 & \emph{6,103} \\
\end{tabular}
\end{center}
\caption{Confusion matrix of some of the worst performing languages by absolute numbers.}
\label{ConfusionThree}
\end{table}

\section{Conclusions and future work}

In the beginning, we were worried about not getting enough new sentences from a simple re-crawl of the old addresses. In the end, the new sentences created an interesting setting for a language identification shared task. The three tracks highlighted different aspects of the problem of language identification.

The next edition of the ULI shared task will incorporate new sentences from the 2018 crawl performed by the SUKI project. Unlike the 2017 edition, the 2018 was a real crawl and much more new material was found.

\section*{Acknowledgements}

\bibliographystyle{coling}
\bibliography{coling2020}

\end{document}